 \pdfoutput=1

\documentclass{article}
\usepackage{ijcai22}

\usepackage{times}

\usepackage{soul}
\usepackage{url}
\usepackage[hidelinks]{hyperref}
\usepackage[utf8]{inputenc}
\usepackage[small]{caption}
\usepackage{graphicx}
\usepackage{amsmath}
\usepackage{booktabs}
\usepackage{multirow}

\title{Image Data Augmentation for Deep Learning: A Survey}

\author{
	Suorong Yang$^{1,2}$\and 
	Weikang Xiao$^{1,3}$\and
	Mengchen Zhang$^3$\and
	Suhan Guo$^{1,3}$\and
	Jian Zhao$^4$\and
	Furao Shen$^{1,2}$\footnote{Contact Author} 
	\affiliations
	$^1$ State Key Laboratory for Novel Software Technology, Nanjing University, China\\
	$^2$ Department of Computer Science and Technology, Nanjing University, China\\
	$^3$School of Artificial Intelligence, Nanjing University, China\\
	$^4$School of Electronic Science and Engineering, Nanjing University, China\\
	\emails
	sryang@smail.nju.edu.cn,
	mg20370042@smail.nju.edu.cn,
	zhangmengchen@pjlab.org.cn,
	dg20370004@smail.nju.edu.cn,
	jianzhao@nju.edu.cn,
	frshen@nju.edu.cn
}

\begin{document}

\maketitle

\begin{abstract}
Deep learning has achieved remarkable results in many computer vision tasks. Deep neural networks typically rely on large amounts of training data to avoid overfitting. However, labeled data for real-world applications may be limited. 
By improving the quantity and diversity of training data, data augmentation has become an inevitable part of deep learning model training with image data.

As an effective way to improve the sufficiency and diversity of training data, data augmentation has become a necessary part of successful application of deep learning models on image data.
In this paper, we systematically review different image data augmentation methods. We propose a taxonomy of reviewed methods and present the strengths and limitations of these methods.
We also conduct extensive experiments with various data augmentation methods on three typical computer vision tasks, including semantic segmentation, image classification and object detection.
Finally, we discuss current challenges faced by data augmentation and future research directions to put forward some useful research guidance.
\end{abstract}
\section{Introduction}

Deep learning has made incredible progress in many fields, including computer vision(CV)~\cite{computer_vision}, recommender system (RS)~\cite{recommender_system}, natural language processing (NLP)~\cite{natural_language_processing}  and so on. The development of these research field 
is mainly affected by the following three aspects: the progress of deep network architectures, great computing power, and access to big data.	  

Firstly, the scale of the network architectures is often proportional to its generalization ability, such as 152-layers ResNet~\cite{resnet}, which, compared with shallow network, can gain significant accuracy form the increased depth.
Secondly, the development of computing power has a significant impact on deep learning. With stronger computing power, it is possible to design models with deeper architecture.
Finally, sufficient open datasets like Imagenet~\cite{imagenet}, MS-COCO~\cite{coco} and PASCAL VOC~\cite{pascal_voc} are crucial to the development of deep learning models. 

However, we observe some imbalance among the developments of these three perspectives.
While various network architectures for different CV tasks have been proposed and the computation power of  graphics processing unit (GPU) have rapidly been increasing, fewer attention has been paid to using data augmentation methods to generate qualified training data. The core idea of data augmentation is to improve the sufficiency and diversity of training data by generating synthetic dataset. The augmented data can be regarded as being extracted from a distribution that is close to the real one. 
Then,  the augmented dataset can represent more comprehensive characteristics. But some research challenges remain in image data augmentation methods. 
First, image data augmentation techniques can be applied into various CV tasks, such as, object detection~\cite{object_detection}, semantic segmentation~\cite{segmentation} and image classification~\cite{image_classification}.	 
But the challenge is that data augmentation methods are tasks-independent. Because the operations are performed on the image data and labels at the same time, and the label types are different under different tasks, the data augmentation methods for object detection task can not be directly applied to semantic segmentation task. This results in inefficiency and low scalability.


Second, there is no theoretical research on data augmentation. For example, there is no quantitative standard on the size of sufficient training datasets. The size of generated training data is usually designed according to personal experience and extensive experiments. 
In addition, paradox may exist when the size of the original dataset is so small. We will face the challenge of how to generate qualified data based on very little data.

To the best of our knowledge, works related to image data augmentation research did not review image data augmentation methods in terms of CV tasks.
One work~\cite{survey1} explores and compares multiple solutions to the problem of data augmentation in image classification, but it only relates to image classification task and experiments with only traditional transformations and GANs.
\cite{face_aug_survey} reviews existing face data augmentation works from perspectives of the transformation types and deep learning. However, the survey is  aimed at the face recognition tasks only.
One work mainly focuses on different data augmentation techniques based on data warping and oversampling~\cite{survey3}. However, it does not provide a systematic review of different approaches.
Another work closely related to ours is~\cite{image_aug_survey} which present some existing methods and promising developments of data augmentation. However, it does not provide an evaluation on the effectiveness of data augmentation for various actual tasks and lacks some newly proposed methods, such as, CutMix~\cite{cutmix}, AugMix~\cite{augmix}, GridMask~\cite{gridmask}, etc.

\begin{figure}[]
	\centering
	\includegraphics[width=0.5\textwidth]{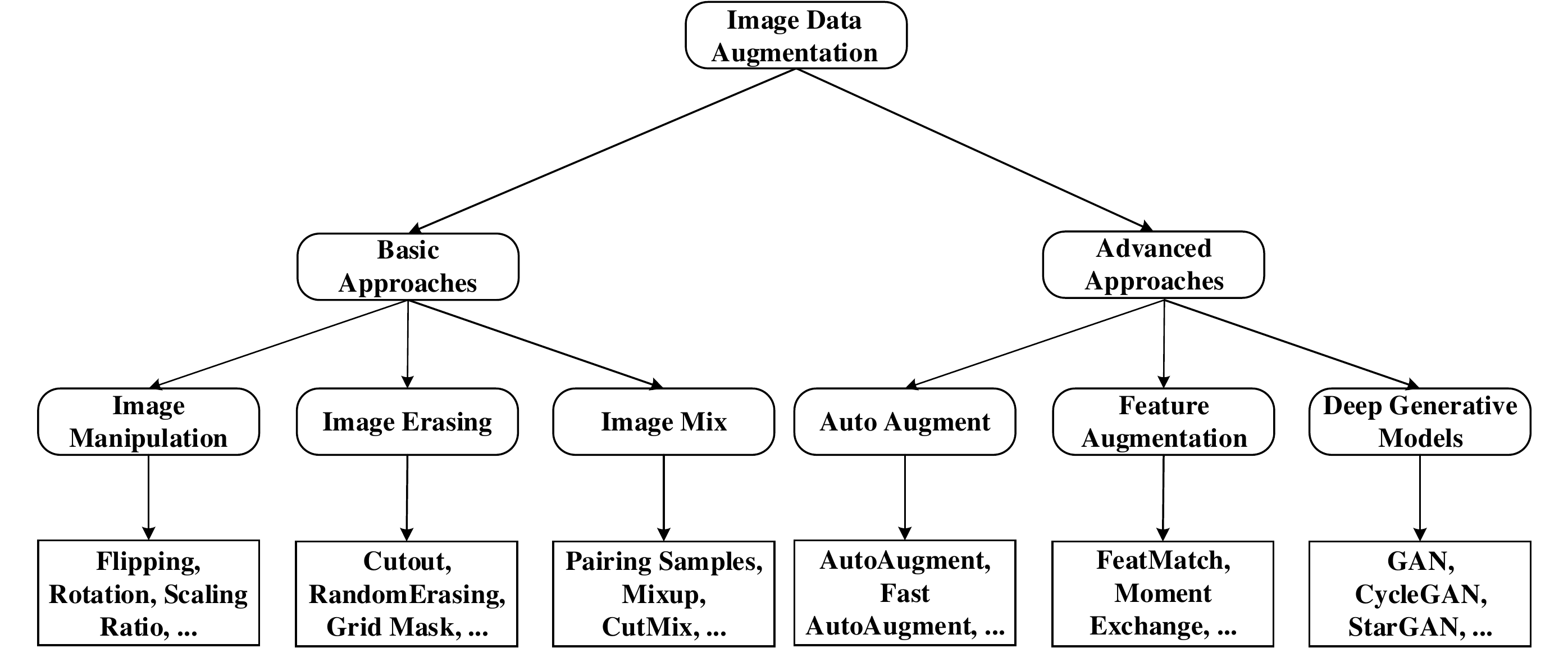}
	\caption{A taxonomy of image data augmentation methods.}
	\label{taxnomy}
\end{figure}

In this paper, we aim to fill the aforementioned gaps by summarizing existing novel image data augmentation methods. To this end, we propose a taxonomy of image data augmentation methods, as illustrated in Fig.~\ref{taxnomy}. Based on this taxonomy, we systematically review the data augmentation techniques from the perspectives of common CV tasks, including object detection, semantic segmentation and image classification. 
Furthermore, we also conduct experiments from the perspectives of these three CV tasks. 
Based on experiment results, we compare the performance of different kinds of data augmentation methods and their combinations on various deep learning models with open image datasets. 
We will also discuss future directions for image data augmentation research.

The reminder of this paper is organized as follows. We present the basic data augmentation methods first, such as traditional image manipulation, image erasing based methods and image mix based methods.
Then we discuss some advanced techniques, including auto augment based methods, feature augmentation techniques, and deep generative models. 
To evaluate the effect of various kinds of data augmentation methods, we conduct experiments in three typical CV tasks with various common public image datasets.
Finally, we highlight some promising directions for future research.

\section{Basic Data Augmentation Methods}
\subsection{Image Manipulation}
Basic image manipulations are focusing on image transformations, such as rotation, flipping, and cropping, etc. Most of these techniques manipulate the images directly and are easy to implement. The methods considered are shown with a concise description in Table.~\ref{basic}.

\setlength{\tabcolsep}{7mm}{
	\begin{table*}[]
		\renewcommand\arraystretch{1.5}
		\caption{Basic image manipulations and concise description.}
		\begin{center}
			\begin{tabular}{c|c}
				\toprule  
				Methods & Description\\
				\midrule  
				Flipping& Flip the image horizontally, vertically, or both. \\
				Rotation& Rotate the image at an angle.\\
				Scaling Ratio& Increase or reduce the image size. \\
				Noise injection& Add noise into the image.\\
				Color space& Change the image color channels.\\
				Contrast& Change the image contrast.\\
				Sharpening& Modify the image sharpness.\\
				Translation& Move the image horizontally, vertically, or both.\\
				Cropping & Crop a sub-region of the image. \\
				\bottomrule 
			\end{tabular}
		\end{center}
		\label{basic}
\end{table*}}

However, the drawbacks exist.
First of all, it is meaningful to apply basic image manipulations only under the assumption that the  existing data obeys the distribution close to the actual data distribution.
Secondly, some basic image manipulation methods, such as translation and rotation, suffer from the padding effect. That is, after the operation, some areas of the images will be moved out of the boundary and lost. Therefore, some interpolation methods will be applied to fill in the blank part. 
Generally, the region outside the image boundary is assumed to be constant 0, which will be black after manipulation.
Moreover, regardless of the CV task, the object of interest should not to be moved off the frame.

\subsection{Image Erasing}
Image augmentation approaches based on image erasing typically delete one or more sub-regions in the image. The main idea is to replace the pixel values of these sub-regions with constant values or random values.  

In~\cite{cutout}, authors considered a simple regularization technique of randomly masking out square regions of input during training convolutional neural networks (CNNs), which is known as cutout. This method is capable of improving the robustness and overall performance of CNNs.
\cite{HaS} proposed Hide-and-Seek (HaS) to hide patches in a training image randomly, which can force the network to seek other relevant content while the most discriminative content is hidden.
~\cite{random_erasing} proposed random erasing, which selects a rectangle region in an image randomly and replaces its pixels with random values. This method is simple, but makes significant improvements.
Recently, in~\cite{gridmask}, authors analyzed the requirement of information dropping and then proposed a structured method, GridMask, which is also based on the deletion of regions in the input images. Unlike Cutout and HaS, GridMask neither removes a continuous region nor randomly selects squares, the deleted regions are a set of spatially uniformly distributed squares, which can be controlled in terms of density and size. Furthermore, to balance the object occlusion and information retention, FenceMask~\cite{fencemask} was proposed, which is based on the simulation of object occlusion strategy. 

\subsection{Image Mix}
Image mix data augmentation has received increasing attention in recent years.
These methods are mainly completed by mixing two or more images or sub-regions of images into one.

In~\cite{pairing_samples}, authors enlarge the dataset by synthesizing every new image with two images randomly selected in the training set, known as pairing samples. The synthesis method used is to average the intensity of two images on each pixel.
~\cite{mixup} discusses a more general synthesis method, Mixup. Mixup, which is not just average the intensity of two images, conducts convex combinations sample pairs and their labels. Therefore, Mixup establishes a linear relationship between data augmentation and the supervision signal and can regularize the neural network to favor simple linear behavior in-between training samples. 
Similar with pairing samples and Mixup, ~\cite{cutmix} proposes the CutMix. Instead of simply removing pixels or mixing images from training set. CutMix replaces the removed regions with a patch from another image and can generate more natural images compared to Mixup. 
~\cite{fmix} proposes Fmix that uses random binary masks obtained by applying a threshold to low-frequency images sampled from Fourier space.
Fmix can take on a wide range of shapes of random masks and can improve performance over Mixup and CutMix. Instead of mixing multiple samples, AugMix~\cite{augmix} first mixes multiple augmentation operations into three augmentation chains and then mixes together the results of several augmentation chains in convex combinations. Therefore, the whole process is typically mixing the results generated by the same image in different augmentation pipelines. In ManifoldMix~\cite{manifold_mix}, authors improve the hidden representations and decision boundaries of neural networks at multiple layers by mixing hidden representations rather than input samples.
\section{Advanced Approaches}
\subsection{Auto Augment}
Instead of manually designing data augmentation methods, researchers try to automatically search augmentation approaches to obtain improved performance. Auto augment has been the frontier of deep learning research and been extensively studied.
Auto augment is based on the fact that different data have different characteristics, so different data augmentation methods have different benefits. Automatic searching for augmentation methods can bring more benefits than manual design.
\cite{autoaugment} describes a simple procedure called AutoAugment to automatically search for improved data augmentation policies. Specifically, AutoAugment consists of two parts: search algorithm and search space. The search algorithm is designed to find the best policy regarding highest validation accuracy. The search space contains many policies which details various augmentation operations and magnitudes with which the operations are applied. 
However, a key challenge of auto augmentation methods is to choose an effective augmentation policy from a large search space of candidate operations. 
The search algorithm usually uses Reinforcement Learning~\cite{reinforcement}, which brings high time cost.
Therefore, to reduce the time cost of AutoAugment, ~\cite{fast_autoaugment} proposes Fast AutoAugment that finds effective augmentation policies via a more efficient search strategy based on density matching. In comparison to AutoAugment, this method can speed up the search time. Meanwhile, ~\cite{PBA} proposes Population Based Augmentation (PBA) to reduce the time cost of AutoAugment which generates nonstationary augmentation policy schedules instead of a fixed augmentation policy. PBA can match the performance of AutoAugment on multiple datasets with less computation time.
Recently, ~\cite{randaugment} proposed RandAugment surpassing all previous automated augmentation techniques, including AutoAugment and PBA.
RandAugment dramatically reduces the search space for data augmentation by removing a separate search, which is computationally expensive. In addition, RandAugment further improves the performance of AutoAugment and PBA. 

However, data augmentation might introduce noisy augmented examples and bring negative influence on inference. Therefore,  ~\cite{keepaugment} proposed KeepAugment to use the saliency map to detect important regions on the original images and then preserve these informative regions during augmentation. KeepAugment automatically improve the data augmentation schemes, such as AutoAugment. In~\cite{aug_improving}, authors observed that the augmentation operations in the later training period are more influential and proposed Augmentation-wise Weight Sharing strategy. Compared with AutoAugment, this work improves efficiency significantly and make it affordable to directly search on large scale datasets. 
Unlike auto-augmentation methods searching strategies in an offline manner,  ~\cite{OHL} formulates the augmentation policy as a parameterized probability distribution and the parameters can be optimized jointly with network parameters, known as OHL-Auto-Aug.
\subsection{Feature Augmentation}
Rather than conduct augmentation only in the input space, feature augmentation performs the transformation in a learned feature space. In~\cite{feature_aug}, authors claimed that when traversing along the manifold it is more likely to encounter realistic samples in feature space than compared to input space. Therefore, various augmentation methods by manipulating the vector representation of data within a learned feature space are investigated, which includes adding noise, nearest neighbor interpolation and extrapolation.
Recently, ~\cite{featmatch} proposed FeatMatch which is a novel learned feature-based refinement and augmentation method to produce a varied set of complex transformations. Moreover, FeatMatch can utilize information from both within-class and across-class prototypical representations.
More recently, authors in~\cite{feat_norm} proposed an implicit data augmentation method, known as Moment Exchange, by encouraging the models to utilize the moment information of latent features. Specifically, the moments of the learned features of one training image are replaced by those of another.
\subsection{Deep Generative Models }
The ultimate goal of data augmentation is to draw samples from the distribution, which represent the generating mechanism of dataset.
Hence, the data distribution we generate data from should not be different with the original one. This is the core idea of deep generative models. Among all the deep generative models methods, generative adversarial networks (GANs)~\cite{gan} are very representative methods. On the one hand, the generator can help generate new images. On the other hand, the discriminator ensures that the gap between the newly generated images and the original images is not too large. Although GAN has indeed been a powerful technique to perform unsupervised generation to augment data~\cite{gan_intro}, how to generate high-quality data and evaluate them still remains a challenging problems. In this subsection, we would like to introduce some image data augmentation techniques based on GAN.

In~\cite{pix2pix}, based on conditional adversarial networks~\cite{cGAN}, authors proposed Pix2Pix to learn the mapping from the input images to output images. However, to train Pix2Pix, a large amount of paired data is needed. It is challenging to collect the paired data. Therefore, in ~\cite{cycleGAN}, unlike Pix2Pix, a CycleGAN model is proposed to learn the translation on an image from the source domain $X$ to a target domain $Y$ in the absence of paired samples.
As the number of source and target domains increases, CycleGAN has to train models for each paired domain separately. For instance, if the task is to do transformation among $n$ domains, we need to train $n \times (n-1)$ models between every two domains. To deal with this issue, ~\cite{stargan} proposed StarGAN to improve the scalability and robustness in handling more than two domains. Generally, StarGAN builds only one model to perform image-to-image translation among multiply domains. In the generation phase, we just need to provide the generator with the source image and an attribute label which indicates the target domain. However, StarGAN takes the domain label as an additional input and learns a deterministic mapping per each domain, which may result in the same output per each domain given an input image. To address this problem, ~\cite{starganv2} proposed StarGAN $v2$, which is a scalable approach that can generate diverse images across multiple domains. 
In this work, researchers define the domain and style of images as visually distinct category groups and the specific appearance of each image, respectively.
For example, the dog can be used as a domain, but there are many kinds of dogs, such as Labrador and Husky. Therefore, the specific dog breed can be viewed as the style of an image. 
In this way, StarGAN $v2$ can translate an image of one domain to diverse images of a target domain, and support multiple domains.
\section{Evaluation}
In this section, based on our taxonomy, we conduct extensive evaluations in three typical CV tasks, semantic segmentation, image classification, and object detection, to show the effectiveness of data augmentation in improving performance. To show fairness,  we use the most and commonly used public datasets for this task. 
\subsection{Semantic Segmentation}
\label{segmentation}
In this subsection, we conduct experiments of semantic segmentation on PASCAL VOC dataset. In table~\ref{semantic_segmentation}, we report the performance improvement on Intersection over Union(IoU) metric with several semantic segmentation models: deeplabv3+~\cite{dplabv3+}, PSPNet~\cite{pspnet}, GCNet~\cite{gcnet}, and ISANet~\cite{isanet}. 

In our experiment, we apply different data augmentation methods based on our taxonomy in Fig~\ref{taxnomy}. Specifically, the applied image manipulation methods include flipping, scaling ratio, rotation, noise injection, cropping, translation and sharpening.
The applied image erasing methods include random erasing, GridMask, FenceMask, Cutout, and HaS.
The applied image mix methods include mosaic, Mixup, CutMix, and Fmix. 
Table~\ref{semantic_segmentation} presents the mean IoU  with and without data augmentation on semantic segmentation models. We observe that data augmentation methods bring IoU improvements for all models.

\begin{table}[h]
	\huge
	\resizebox{.95\columnwidth}{!}{
		\renewcommand\arraystretch{1.5}{
		\begin{tabular}{llll}
			\toprule[2pt]
			Model   & {\begin{tabular}[c]{@{}l@{}}w/o aug\\ (\%)\end{tabular}} &{\begin{tabular}[c]{@{}l@{}}w/ aug\\ (\%)\end{tabular}}& {\begin{tabular}[c]{@{}l@{}}Improvement\\ (\%)\end{tabular}} \\ \midrule
			DeepLabV3+ &75.32\% &75.75\% &0.53\% \\
			PSPNet     &73.38\% &74.33\% &0.95\% \\
			GCNet      &71.86\% &72.93\% &1.07\% \\ 
			ISANet	   &71.65\% &74.26\% &2.61\% \\
			\bottomrule[1.5pt]
	\end{tabular}}}
	\caption{Semantic segmentation improvement from data augmentation based on IoU and accuracy.}
	\label{semantic_segmentation}
	
\end{table}
\subsection{Image Classification}
In this experiment, we compare the classification accuracy with and without augmentation on several widely used image classification techniques, including  Wide-ResNet~\cite{wide_resnet}, DenseNet~\cite{densenet}, and Shake ResNet~\cite{shake}. These models are evaluated on several public image classification datasets, including CIFAR-10~\cite{cifar-10}, CIFAR-100~\cite{cifar-100} and SVHN~\cite{SVHN}.  Moreover, the data augmentation methods applied are the same with those in~\ref{segmentation}, which includes several image manipulation methods, image erasing methods and image mix methods.

Table~\ref{classify_res} summarizes the image classification results with and without data augmentation. It can be observed that data augmentation leads to average accuracy improvement(AAI).
\begin{table}[ht]
 \LARGE
		\resizebox{.95\columnwidth}{!}{
			 \renewcommand\arraystretch{1.5}{
			\begin{tabular}{lllll}
				\toprule[1.5pt]
				Dataset  & Model & {\begin{tabular}[c]{@{}l@{}}w/o aug\\ (\%)\end{tabular}} &{\begin{tabular}[c]{@{}l@{}}w/ aug\\ (\%)\end{tabular}}& {\begin{tabular}[c]{@{}l@{}}AAI\\ (\%)\end{tabular}} \\ \midrule
				& DenseNet & 94.15 & 94.59 & 0.44  \\  
				CIFAR-10 & Wide-ResNet & 93.34 & 94.67 &  1.33    \\ 
				& Shake-ResNet &93.7 & 94.84 & 1.11    \\ \midrule 
				& DenseNet & 74.98  & 75.93  & 0.95   \\  
				CIFAR-100 & Wide-ResNet & 74.46  &  76.52  & 2.06    \\ 
				& Shake-ResNet & 73.96   & 76.76  &  2.80  \\  \midrule
				& DenseNet & 97.91  & 97.98  & 0.07   \\  
				SVHN & Wide-ResNet & 98.23  &  98.31  & 0.80    \\ 
				& Shake-ResNet & 98.37  & 98.40  & 0.30     \\  
				\bottomrule[1.5pt]
		\end{tabular}}}
		\caption{Image classification accuracy improvement from data augmentation on CIFAR-10, CIFAR-100, and SVHN.}
		\label{classify_res}
	\end{table}
 
	\subsection{Object Detection}
	In this subsection, we compare the effectiveness of various image data augmentation methods on widely used COCO2017 dataset, which is usually used for object detection task. We demonstrate the experiment results with and without data augmentation in two popular object detection deep models, FasterRCNN~\cite{faster-rcnn} and CenterNet~\cite{centernet}. We consider the data augmentation methods used are the same with those in~\ref{segmentation}, which includes several image manipulation methods, image erasing methods and image mix methods.  In table~\ref{det_res}, we report the performance improvement on mean average precision (mAP), AP50 and AP75 and we summarize these metrics for all methods in average sense.
	We observe that the data augmentation methods bring promising performance improvements. 
                                                          
	\begin{table}[ht]
		\LARGE
			\resizebox{.95\columnwidth}{!}{
				\begin{tabular}{llll}
					\toprule[1.5pt]
					Metric  &   &Faster R-CNN & CenterNet \\ \midrule
					& w/o aug & 36.40 & 41.42    \\  
					{\begin{tabular}[c]{@{}l@{}}mAP\\ (\%)\end{tabular}}  & w/ aug & 36.80 & 41.15      \\ 
					& API & 2.40 & -0.27      \\ \midrule 
					& w/o aug & 57.20 & 58.29   \\  
					{\begin{tabular}[c]{@{}l@{}}AP50\\ (\%)\end{tabular}} & w/ aug & 58.0 &  58.01   \\ 
					& API & 0.80 & -0.28    \\  \midrule
					& w/o aug & 39.50 & 45.53    \\  
					{\begin{tabular}[c]{@{}l@{}}AP75\\ (\%)\end{tabular}} & w/ aug & 40.0 &  45.30    \\ 
					& API & 0.50 & -0.23    \\  
					\bottomrule[1.5pt]
			\end{tabular}}
			\caption{Results of object detection on COCO2017 dataset with and without data augmentation methods applied.}
			\label{det_res}
		\end{table}
	\section{Discussion for Future Directions}
	Despite extensive efforts on image data augmentation research to bring performance improvement on deep learning models, several open problems remain yet to completely to solve, which are summarized as follows.\\
	\textbf{Theoretical Research on Data Augmentation.} There is a lack of theoretical research on data augmentation. Data augmentation is more regarded as an auxiliary tool to improve the performance. Specifically, some methods can improve the accuracy, but we do not fully understand the reasons behind, such as pairing samples and mixup. To human eyes, the augmented data with pairing samples and mixup are visually meaningless. Furthermore, there is no theory on the size of sufficient training datasets.  The size of the dataset suitable for tasks and models is usually designed based on personal experience and through extensive experiments. For example, researchers determine the size of datasets according to the specific models, training objectives, and the difficulty of data collection.  	
	Rigorous and thorough interpretability can not only explain why some augmentation techniques are useful but also can help guide the process of choosing or designing the most applicable and effective methods to enlarge our datasets. Thus, a critical future perspective is to develop theoretical support for data augmentation.\\
	\textbf{The Evaluation of Data Augmentation Methods.} The quantity and diversity of training data are of great importance to model's generalization ability. However, since there is no unified metrics,	how to evaluate the synthesized image quality is an open problem~\cite{evaluate_DA}. At this stage, researchers evaluate the quality of the synthetic data in several following ways. 
	First, the synthetic data are usually evaluated by human eyes, which is time-consuming, labor-intensive, and subjective. Amazon Mechanical Turk (AMT) is often used to evaluate the realism of outputs. AMT evaluates the quality and authenticity of the generated images by asking participants to vote for various images synthesized with different methods. 
	Second, some studies combine the evaluation with specific tasks, which is to evaluate data augmentation methods according to their effect on the tasks metrics with and without data augmentation, such as classification tasks with classification accuracy and semantic segmentation with IOU of masks.
	However, there is no evaluation index only for the synthetic data itself. 
	Generally, the evaluation metrics are based on diversity of individual data and consistency of overall data distribution regardless of what task is. Data quality analysis can help design evaluation metrics .\\
	\textbf{Class Imbalance.} class imbalance or very few data can  severely skew the data distribution~\cite{data_imbalan}. This situation occurs since the learning process is often biased toward the majority class examples, so that minority ones are not well modeled.
	 Synthetic minority of oversampling technique (SMOTE)~\cite{smote} is to oversample the minority class. However, the oversampling is repeating drawing from the current data set.
	This may saturate the minority class and cause overfitting. Ultimately, we expect generated data can simulate distribution similar with training data while diversity never losses.\\
	 \textbf{The Number of Generated Data.} 	An interesting point for data augmentation is that the increase in the amount of training data is not exactly proportional to the increase in the performance. When a certain amount of data is reached, continue to increase the data without improving the effect. This may partly because, despite the increase in the number of data, the diversity of data remains unchanged.
	 Thus, how much data should be generated is good enough to improve the model performance remains to be further explored.\\
	\textbf{The Selection and Combination of Data Augmentation.} Since various data augmentation can be combined together to generate new image data, the selection and combination of data augmentation techniques are critical.  Image recognition experiment shows that results from~\cite{DA_experiment} combined methods are often better than single method. Therefore, how to choose and combine methods is a key point when performing data augmentation.
	However, from our evaluation, the methods applicable for different datasets and tasks are not the same.
	 Therefore, the set of augmentation methods must be carefully designed, implemented, and tested for every new task and dataset. 
	 
	\section{Conclusion}
	 With the development of deep learning, the requirements for training datasets are becoming increasingly stringent. Thus we emphasize that data augmentation is an effective solution for the shortage of limited labeled image data.
	 In this paper, we present a comprehensive review on image data augmentation methods in various CV tasks. We propose a taxonomy, summarizing representative approaches in each category. We then compare the methods empirically in various CV tasks. Finally, we discuss the challenges and highlight future perspectives.
 
	\bibliographystyle{named}

\end{document}